\documentclass[12pt]{article}
\usepackage{amssymb}
\usepackage{graphicx}
\usepackage{amsmath}
\usepackage{amsfonts}
\usepackage{amssymb}
\usepackage{subfigure}
\usepackage{overpic}
\usepackage{epsfig}
\usepackage[linesnumbered,vlined,slide,ruled]{algorithm2e}

\newcommand{\bb}[1]{\boldsymbol{\mathrm{#1}}}

\def\Tr{\mathrm{T}}

\newcommand{\EE}{\mathbb{E}}
\newcommand{\RR}{\mathbb{R}}
\newcommand{\HH}{\mathbb{H}}

\newcommand{\xx}{\bb{x}}
\newcommand{\yy}{\bb{y}}

\newcommand{\ppp}{\bb{p}}
\newcommand{\qqq}{\bb{q}}

\newcommand{\sign}{\mathrm{sign}}

\newcommand{\uu}{\bb{u}}
\newcommand{\vv}{\bb{v}}

\newcommand{\bbb}{\bb{b}}
\newcommand{\aaa}{\bb{a}}

\newcommand{\Pp}{\bb{P}}
\newcommand{\Qq}{\bb{Q}}

\newcommand{\Kk}{\bb{K}}
\newcommand{\Aa}{\bb{A}}
\newcommand{\Bb}{\bb{B}}

\newcommand{\Vv}{\bb{V}}
\newcommand{\Uu}{\bb{U}}

\newcommand{\Ss}{\bb{S}}
\newcommand{\Ii}{\bb{I}}

\newcommand{\SSigma}{\bb{\Sigma}}
\newcommand{\aalpha}{\bb{\alpha}}
\newcommand{\bbeta}{\bb{\beta}}

\newcommand\tr{\mathrm{tr}\,}

 {\everymath{\displaystyle\everymath{}}\array}%
 {\endarray}
\begin{document}

\title{Multimodal diff-hash}

\author{
Michael M. Bronstein\\
{\small Institute of Computational Science, Faculty of Informatics,}\\ 
{\small Universit{\`a} della Svizzera Italiana}\\ 
{\small Via G. Buffi 13, Lugano 6900, Switzerland}\\
{\tt\small michael.bronstein@usi.ch}\\
}

\maketitle

\begin{abstract}
Many applications require comparing multimodal data with different structure and dimensionality that cannot be compared directly. Recently, there has been increasing interest in methods for learning and efficiently representing such multimodal similarity. 
In this paper, we present a simple algorithm for multimodal similarity-preserving hashing, trying to map multimodal data into the Hamming space while preserving the intra- and inter-modal similarities. We show that our method significantly outperforms the state-of-the-art method in the field.  
\end{abstract}

\section{Introduction}

The need to model and compute similarity between some objects is central to many applications ranging from medical imaging to biometric security. In various problems in different fields we need to compare object as different as functions, images, geometric shapes, probability distributions, or text documents. Each such problem has its own notion of data similarity. 

%Of particular interest are the fields of image sciences, computer vision, and pattern recognition, which have recently experienced a huge challenge coming from the explosive growth of the amount of visual and geometric data (images, videos, 3D objects) that are being created and made available in the public domain \cite{videodna}.

A particularly challenging case of similarity arises in applications dealing with multimodal data, which have different representation, dimensionality, and structure. Data of this kind is encountered prominently in medical imaging (e.g. fusion of different imaging modalities like PET and CT) \cite{isbi} and multimedia retrieval (e.g. querying image databases by text keywords) \cite{shapegoogle}. Such data are incomparable as apples and oranges by means of standard metrics and require the notion of multimodal similarity. 

While such multimodal similarity is difficult to model, in many cases it is easy to learn from examples. For instance, in Internet vision applications we can easily obtain multiple examples of visual objects with a binary similarity function telling whether two objects are similar or not. 
Learning and representing such similarities in a convenient way is a big challenge. 

Particular setting of similarity representation problem is similarity sensitive hashing \cite{shakhnarovich}, which has attracted significant attention in the computer vision and pattern recognition communities. 
In \cite{cmssh}, we extended the boosting-based similarity-sensitive hashing (SSH) method to the multimodal setting (referred to as cross-modality SSH or CM-SSH). This is, to the best of our knowledge, the first and the only multimodal similarity-preserving hashing algorithm in the literature.  

The purpose of this paper is to develop a different simpler and efficient multimodal hashing algorithm. 
The rest of the paper is organized as follows. In Section~2, we formulate the problem of multimodal hashing. In Section~3, we overview the CM-SSH algorithm. In Section~4, we propose our new method (cross-modality diff-hash or CM-DIF) and in Section~5 discuss its extension (multimodal kernel diff-hash or MM-kDIF) using kernelization. Section~6 shows some experimental results. 

\section{Background}

Let $X \subseteq \RR^n$ and $Y \subseteq \RR^{n'}$ be two spaces representing data belonging to different modalities (e.g., $X$ are images and $Y$ are text descriptions). 
Note that even though we assume that the data can be represented in the Euclidean space, the similarity of the data is not necessarily Euclidean and in general can be described by some metrics $d_X : X\times X \rightarrow \RR_+$ and $d_Y : Y\times Y \rightarrow \RR_+$, to which we refer as {\em intra-modal dissimilarities}. 
Furthermore, we assume that there exists some {\em inter-modal dissimilarity} $d_{XY}: X\times Y \rightarrow \RR_+$ quantifying the ``distance'' between points in different modality. 
The ensemble of intra- and inter-modal structures $d_X, d_Y, d_{XY}$ is not necessarily a metric in the strict sense. In order to deal with these structures in a more convenient way, we try to represent them in a common metric space.

%\subsection{Multimodal hashing problem}

The broader problem of {\em multimodal hashing} is to represent the data from different modalities $X, Y$ in a common space $\HH^m = \{ \pm 1\}^m$ of $m$-dimensional binary vectors with the Hamming metric $d_{\HH^m}(a, b) =  \frac{m}{2} - \frac{1}{2} \sum_{i=1}^m a_i b_i$ by means of two embeddings, $\xi: X \rightarrow \HH^m$ and $\eta : Y \rightarrow \HH^m$ mapping similar points as close as possible to each other and dissimilar points as distant as possible from each other, such that $d_{\HH^m} \circ (\xi \times \xi) \approx d_{X}$, $d_{\HH^m} \circ (\eta \times \eta) \approx d_{Y}$, and $d_{\HH^m} \circ (\xi \times \eta) \approx d_{XY}$. 
In a sense, the embeddings act as a {\em metric coupling}, trying to construct a single metric $d_{\HH^m} \circ (\xi \times \eta)$ that preserves the intra- and inter-modal similarities.

%\subsection{Cross-modality hashing problem}

A simplified setting of the multimodal hashing problem is {\em cross-modality hashing}, in which only the inter-modal dissimilarity $d_{XY}$ is taken into consideration and $d_X, d_Y$ are ignored.

For simplicity, in the following discussion we assume the inter-modal dissimilarity to be binary, $d_{XY} \in \{0, 1\}$, i.e., a pair of points can be either similar or dissimilar. This dissimilarity is usually unknown and hard to model, however, it should be possible to sample $d_{XY}$ on some subset of the data $X' \subset X, Y' \subset Y$.  
This sample can be represented as set of similar pairs of points ({\em positives}) $\mathcal{P} = \{ (x \in X', y\in Y') : d_{XY}(x,y) = 0\}$ and a set of dissimilar pairs of points ({\em negatives}) $\mathcal{N} = \{ (x \in X', y\in Y') : d_{XY}(x,y) = 1\}$.

The problem of cross-modality hashing thus boils down to find two embeddings $\xi: X \rightarrow \HH^m$ and $\eta : Y \rightarrow \HH^m$ such that $m\, d_{\HH^m} \circ (\xi \times \eta) \approx d_{XY}$. 
Alternatively, this can be expressed as having $\EE \{ d_{\HH^m} \circ (\xi \times \eta) | \mathcal {P} \}\approx 0$ (i.e., the hash has high collision probability on the set of positives) and $\EE \{ d_{\HH^m} \circ (\xi \times \eta) | \mathcal {N} \} \gg 0$. 
The former can be interpreted as the {\em false negative rate} (FNR) and the latter as the {\em false positive rate} (FPR).

\section{Cross-modality similarity-sensitive hashing (CM-SSH)}

To further simplify the problem, consider embeddings given in parametric form as $\xi(\xx) = \sign(\Pp \xx + \aaa)$ and $\eta(\yy) = \sign(\Qq \yy + \bbb)$ \cite{shakhnarovich,cmssh}. Here, $\Pp, \Qq$ are {\em projection} matrices of size $m\times n$ and $m\times n'$, respectively, and $\aaa, \bbb$ are {\em threshold} vectors of size $m\times 1$.

In \cite{cmssh}, we introduced the {\em cross-modality similarity-sensitive hashing} (CM-SSH) method, which is to the best of our knowledge, the first and the only multimodal hashing algorithm existing to date. 
The idea closely follows the similarity-sensitive hashing (SSH) method \cite{shakhnarovich}, considering the hash construction as boosted binary classification, where each hash dimension acts as a weak binary classifier. % 
For each dimension, AdaBoost is used to maximize the following loss function
\begin{eqnarray}
\label{eq:mm_loss-ssh}
\min_{\ppp_i,\qqq_i, a_i, b_i}  \,\,\, \sum_{(\xx,\yy) \in \mathcal{P} \cup \mathcal{N}} w_{i}(\xx,\yy) s(\xx,\yy) \sign(\ppp_i^\Tr \xx + a_i) \, \sign(\qqq_i^\Tr \yy + b_i), 
\end{eqnarray}
where $s(\xx, \yy) = 1 - 2d_{XY}(\xx, \yy)$ is binary intra-modal similarity and $w_{i}(\xx,\yy)$ is the AdaBoost weigh for pair $(\xx,\yy)$ at $i$th iteration. 
Since the minimization problem~(\ref{eq:mm_loss-ssh}) is difficult, it is relaxed in the following way \cite{cmssh}: First, removing the non-linearity and setting $a_i = b_i = 0$, find the projection vectors $\ppp_i, \qqq_i$. Then, fixing the projections $\ppp_i, \qqq_i$, find the thresholds $a_i, b_i$. 
%
%For the first step, Shakhnarovich {\em et al.} \cite{shakhnarovich} used random projections in the SSH. A somewhat better way was used in \cite{videodna} in the unimodal setting and in \cite{cmssh} in the multimodal setting, considering the problem 
%%
%\begin{eqnarray*}
%\label{eq:mm_loss-ssh1}
% & \displaystyle\min_{\ppp_i,\qqq_i}&   \sum_{(\xx,\yy) \in \mathcal{P} \cup \mathcal{N}} w_{i}(\xx,\yy) s(\xx,\yy) (\ppp_i^\Tr \xx ) \, (\qqq_i^\Tr \yy) = \\
%& \displaystyle\min_{\ppp_i,\qqq_i} &  \ppp_i^\Tr \left( \sum_{(\xx,\yy) \in \mathcal{P} \cup \mathcal{N}} w_{i}(\xx,\yy) s(\xx,\yy) \xx \yy^\Tr \right) \qqq_i = \displaystyle\min_{\ppp_i,\qqq_i}  \ppp_i^\Tr \Cc \qqq_i. 
%\end{eqnarray*}
%%
%The problem was solved in \cite{cmssh} by search in a low-dimensional subspace spanned by the largest left and right singular vectors of the $n\times n'$ matrix $\Cc$. 

The disadvantages of the boosting-based CM-SSH is first high computational complexity, and second, the tendency to find unnecessary long hashes (the second problem can be partially resolved by using sequential probability testing \cite{waldhash} which creates hashes of minimum expected length).

%The main contribution of this paper is a simpler and better multimodal hashing algorithm. 
%In this paper, we present a simpler and better method 

\section{Cross-modality diff-hash (CM-DIF)}

In \cite{ldahash}, we proposed a different and simpler approach (dubbed {\em diff-hash}) to create similarity-sensitive hash functions in the unimodal setting. We adopt similar ideas here to develop multimodal similarity-sensitive hashing algorithms.

The optimal cross-modality hashing can be found by minimizing the loss 
\begin{eqnarray}
\label{eq:mm_loss-lda}
L & = & \gamma \EE \{ d_{\HH^m} \circ (\xi \times \eta) | \mathcal {P} \} -  \EE \{ d_{\HH^m} \circ (\xi \times \eta) | \mathcal {N} \} \nonumber \\
&=& {\textstyle \frac{m(\gamma - 1)}{2} }+ {\textstyle\frac{1}{2}} \EE \{ \xi^\Tr \eta | \mathcal{N} \} - {\textstyle\frac{\gamma}{2}} \EE \{ \xi^\Tr \eta | \mathcal{P} \}
\end{eqnarray}
with respect to the embedding functions $\xi, \eta$, which is, up to constants, equivalent to minimizing the correlations 
\begin{eqnarray}
\label{eq:mm_loss-lda1}
L(\Pp,\Qq,\aaa,\bbb) & = &  \EE \{ \sign(\Pp \xx + \aaa)^\Tr\sign(\Qq \yy + \bbb) | \mathcal{N} \} \nonumber \\
&-& \gamma \EE \{ \sign(\Pp \xx + \aaa)^\Tr\sign(\Qq \yy + \bbb) | \mathcal{P} \} 
\end{eqnarray}
w.r.t. the projection matrices $\Pp,\Qq$ and threshold vectors $\aaa,\bbb$. 
The first and second terms in (\ref{eq:mm_loss-lda1}) can be thought of as FPR and FNR, respectively. The parameter $\gamma$ controls the tradeoff between FPR and FNR. The limit case $\gamma \gg 1$ effectively considers only the positive pairs ignoring the negative set. %This is helpful in case of label noise. 

Problem (\ref{eq:mm_loss-lda1}) is a highly non-convex non-linear optimization problem difficult to solve straightforwardly. 
Similarly to \cite{ldahash,cmssh}, we simplify the problem in the following way. First, we ignore the threshold and solve a simplified problem without the sign non-linearity for projection matrices, 
\begin{eqnarray*}
\min_{\Pp,\Qq} \,\,\, \EE \{ (\Pp \xx )^\Tr (\Qq \yy ) | \mathcal{N} \} - \gamma \EE \{ (\Pp \xx )^\Tr (\Qq \yy ) | \mathcal{P} \} 
\,\,\, \mathrm{s.t.} \,\,\, \Pp^\Tr \Pp = \Ii_{n},  \,\, \Qq^\Tr \Qq = \Ii_{n'}. 
\end{eqnarray*}
 Second, fixing the projections we find optimal thresholds, 
\begin{eqnarray*}
\min_{\aaa,\bbb} \,\,\, \EE \{ \sign(\Pp \xx + \aaa)^\Tr\sign(\Qq \yy + \bbb) | \mathcal{N} \} - \gamma \EE \{ \sign(\Pp \xx + \aaa)^\Tr\sign(\Qq \yy + \bbb) | \mathcal{P} \}. 
\end{eqnarray*} 
We details each step in Sections~\ref{sec:proj}--\ref{sec:thresh}. The whole method is summarized in Algorithm~\ref{alg:1}.

\subsection{Projection computation}
\label{sec:proj}

Dropping the sign function and the offset, the loss function (\ref{eq:mm_loss-lda1}) becomes 
\begin{eqnarray}
\label{eq:mm_loss-lda2}
L(\Pp,\Qq,\aaa,\bbb) \approx \hat{L}(\Pp,\Qq) & =&  \EE \{ (\Pp \xx )^\Tr (\Qq \yy ) | \mathcal{N} \} - \gamma \EE \{ (\Pp \xx )^\Tr (\Qq \yy ) | \mathcal{P} \}\nonumber\\ 
& = & \tr ( \Pp \EE\{ \xx \yy^\Tr | \mathcal{N}\} \Qq^\Tr) - \gamma \tr ( \Pp \EE\{ \xx \yy^\Tr  | \mathcal{P} \} \Qq^\Tr)\nonumber\\
& = &\tr ( \Pp  (\SSigma_{XY}^{\mathcal{N}}  - \gamma \SSigma_{XY}^{\mathcal{P}}) \Qq^\Tr) = \tr ( \Pp  \SSigma_{XY}^{D} \Qq^\Tr)
\end{eqnarray}
where $\SSigma_{XY}^\mathcal{P}, \SSigma_{XY}^\mathcal{N}$ denote the $n \times n'$ covariance matrices of the positive and negative multi-modal data, respectively, and $\SSigma^D_{XY}$ is the weighted difference of these covariances. 
The name of the algorithm, {\em cross-modality diff-hash} (CM-DIF), refers in fact to this covariance difference matrix. 
Note that in order to avoid trivial solution, we must constrain the projection matrices to be unitary, i.e., $\Pp^\Tr \Pp = \Ii_{n}$ and $\Qq^\Tr \Qq = \Ii_{n'}$.

The difference of covariance matrices has a singular value decomposition of the form $\SSigma^D_{XY} = \Uu \Ss \Vv^\Tr$, where $\Uu$ and $\Vv$ are unitary matrices of singular vectors  of size $n \times n$ and $n'\times n'$, respectively ($\Uu^\Tr \Uu = \Ii_n$, $\Vv^\Tr \Vv = \Ii_{n'}$), and $\Ss$ is a diagonal matrix of singular values of size $n \times n'$. 

It can be easily shown that the loss $\hat{L}(\Qq,\Pp)$ is minimized by setting the projection matrices to be the smallest left and right singular vectors of the matrix $ \SSigma^D_{XY}$, respectively: $\Pp = [\uu_{n-m+1} \hdots \uu_{n}]^\Tr$ and $\Qq = [\vv_{n'-m+1} \hdots \vv_{n'}]^\Tr$. 
From this result it also follows that the problem is separable, and each dimension can be treated independently.

\subsection{Threshold selection}
\label{sec:thresh}

Having the projection matrices $\Pp, \Qq$ fixed, the loss function (\ref{eq:mm_loss-lda1}) can be written as 
\begin{eqnarray}
\label{eq:mm_loss}
L(\aaa,\bbb) & = & \EE \{ \sign(\Pp \xx + \aaa)^\Tr\sign(\Qq \yy + \bbb) | \mathcal {N} \} \nonumber \\
&-&  \gamma \EE \{ \sign(\Pp \xx + \aaa)^\Tr\sign(\Qq \yy + \bbb) | \mathcal {P} \}  \nonumber \\
&=& \textstyle\sum_{i=1}^m\EE \{ \sign(\ppp_i^\Tr \xx + a_i) \sign(\qqq_i^{\Tr} \yy + b_i) | \mathcal {N} \} \nonumber \\
&-&  \gamma \textstyle\sum_{i=1}^m \EE \{ \sign(\ppp_i^\Tr \xx + a_i) \sign(\qqq_i^{\Tr} \yy + b_i) | \mathcal {P} \} 
\end{eqnarray}
The problem is separable and can be solved independently in each dimension $i$. 
%In the following, we omit the dimension index and denote $x' = \ppp_i^\Tr \xx $ and $y' = \qqq_i^\Tr \yy $. Then, 
We express the false positive and negative rates as a function of the thresholds $a_i,b_i$ as 
\begin{eqnarray}
\label{eq:prob1}
\mathrm{FN}_i(a_i,b_i) &=& \mathrm{Pr}(\ppp_i^\Tr \xx  + a_i < 0 | \mathcal{P}) \cdot  \mathrm{Pr}( \qqq_i^\Tr \yy + b_i > 0 | \mathcal{P}) \nonumber \\
&+& \mathrm{Pr}(\ppp_i^\Tr \xx  + a_i > 0 | \mathcal{P}) \cdot  \mathrm{Pr}( \qqq_i^\Tr \yy + b_i < 0 | \mathcal{P}) \nonumber \\
& = & \mathrm{Pr}(\ppp_i^\Tr \xx   < -a_i | \mathcal{P}) \cdot (1 - \mathrm{Pr}( \qqq_i^\Tr \yy < -b_i | \mathcal{P})) \nonumber \\
&+& \mathrm{Pr}(\qqq_i^\Tr \yy  < -b_i | \mathcal{P}) \cdot (1 - \mathrm{Pr}( \ppp_i^\Tr \xx  < -a_i | \mathcal{P})) 
\end{eqnarray}
and
\begin{eqnarray}
\label{eq:prob2}
\mathrm{FP}_i(a_i,b_i) &=& \mathrm{Pr}(\ppp_i^\Tr \xx + a_i < 0 | \mathcal{N}) \cdot  \mathrm{Pr}( \qqq_i^\Tr \yy + b_i < 0 | \mathcal{N}) \nonumber \\
&+& \mathrm{Pr}(\ppp_i^\Tr \xx + a_i > 0 | \mathcal{N}) \cdot  \mathrm{Pr}( \qqq_i^\Tr \yy + b_i > 0 | \mathcal{N}) \nonumber \\
& = & \mathrm{Pr}(\ppp_i^\Tr \xx  < -a_i | \mathcal{N}) \cdot \mathrm{Pr}( \qqq_i^\Tr \yy < -b_i | \mathcal{N}) \nonumber \\
&+& \mathrm{Pr}(\qqq_i^\Tr \yy  < -b_i | \mathcal{N}) \cdot \mathrm{Pr}( \ppp_i^\Tr \xx < -a_i | \mathcal{N}).
\end{eqnarray}
The above probabilities %$\mathrm{Pr}(\ppp_i^\Tr \xx < \cdot ), \mathrm{Pr}(\qqq_i^\Tr \yy < \cdot )$ 
can be estimated from histograms (cumulative distributions) of $ \ppp_i^\Tr \xx$ and $ \qqq_i^\Tr \yy$ on the positive and negative sets. 
Optimal thresholds 
\begin{eqnarray}
\label{eq:opt_thr}
(a_i^*,b_i^*) &=& \mathop{\mathrm{argmin}}_{a,b} \,\, \gamma \mathrm{FN}_i(a,b) + \mathrm{FP}_i(a,b)
\end{eqnarray}
are obtained by means of exhaustive search. 
To reduce the complexity of this search, we define a set of grids on the threshold parameter space.

\begin{algorithm}[t!]
%\linesnumbered
\KwIn{Positive and negative sets $\mathcal{P}, \mathcal{N} \subset X\times Y$ of multimodal data of dimensionality $n$ and $n'$; 
Dimensionality of the hash $m$; Tradeoff parameter $\gamma$.}
\KwOut{Optimal projection matrices $\Pp, \Qq$ of size $m\times n, m\times n'$; optimal offset vectors $\aaa, \bbb$ of size $m\times 1$.}

Compute the $n\times n'$ covariance matrices $\SSigma^\mathcal{P}_{XY}, \SSigma^\mathcal{N}_{XY}$.

Compute the covariance difference matrix $\SSigma^D_{XY} = \SSigma^\mathcal{N}_{XY} - \gamma \SSigma^\mathcal{P}_{XY}$.

Perform singular value decomposition  $\SSigma^D_{XY} = \Uu \Ss \Vv^\Tr$. 

\For{$i=1,\dots,m$}{

Set the $i$th row of the projection matrices to be the $i$th smallest left and right singular vectors, $\ppp^\Tr_i = \uu^\Tr_{n-i+1}$ and $\qqq^\Tr_i = \vv_{n'-i+1}^\Tr$.

Compute the probabilities $\mathrm{Pr}(\ppp_i^\Tr \xx < \cdot | \mathcal{P})$, $\mathrm{Pr}(\qqq_i^\Tr \yy < \cdot  | \mathcal{P})$, 
 $\mathrm{Pr}(\ppp_i^\Tr \xx < \cdot | \mathcal{N})$, $\mathrm{Pr}(\qqq_i^\Tr \yy < \cdot  | \mathcal{N})$. 

Compute the rates $\mathrm{FP}_i(a_i,b_i), \mathrm{FN}_i(a_i,b_i)$ as a function of the thresholds $a_i, b_i$ according to (\ref{eq:prob1})-(\ref{eq:prob2}).

Compute the optimal thresholds 
\begin{eqnarray*}
(a_i^*,b_i^*) &=& \mathop{\mathrm{argmin}}_{a,b} \,\, \gamma \mathrm{FN}_i(a,b) + \mathrm{FP}_i(a,b).
\end{eqnarray*}

}

\caption{Cross-modality diff-hash algorithm (CM-DIF). \label{alg:1}}
\end{algorithm}

\subsection{Hash function application}

Once the projections $\Pp, \Qq$ and thresholds $\aaa,\bbb$ are computed, given new data points $\xx, \yy$, we construct the corresponding $m$-dimensional binary hash vectors as $\xi(\xx) = \sign(\Pp \xx + \aaa)$ and $\eta(\yy) = \sign(\Qq \yy + \bbb)$.

\section{Multimodal kernel diff-hash (MM-kDIF)}

An obvious disadvantage of diff-hash (and spectral methods in general) compared to AdaBoost-based methods is that it must be {\em dimensionality-reducing}: since we compute projections $\Pp$ and $\Qq$ as the singular vectors of a covariance matrix of size $n\times n'$, the dimensionality of the embedding space must satisfy $m \leq \min\{n,n' \}$. 
In some cases, such a dimensionality may be too low and would not allow to correctly separate the data. 
A second disadvantage is of the cross-modality hashing problem in general, that it considers only the inter-modal similarity $d_{XY}$, ignoring the intra-modal similarities $d_X, d_Y$. 

A standard way to cope with the first problem is the {\em kernel trick} \cite{kernel}, which transforms the data into some feature space that is never dealt with explicitly (only inner products in this space, referred to as {\em kernel}, are required). A kernel version of the uni-modal diff-hash was described in \cite{kdiffh}. Here, we show that the use of kernels also allows incorporating intra-modal similarities into the problem.

%In order to simplify the following discussion, we consider the case $\gamma \gg 1$, which simplifies the loss (\ref{eq:mm_loss-lda}) to the set of positives only, $L(\Pp,\Qq,\aaa,\bbb) = \EE \{ \sign(\Pp \xx + \aaa)^\Tr\sign(\Qq \yy + \bbb) | \mathcal{P} \}$. We follow the same relaxation of the optimization problem with first optimizing an approximate loss $L(\Pp,\Qq,\aaa,\bbb) \approx  \hat{L}(\Pp,\Qq) = - \EE \{ (\Pp \xx )^\Tr (\Qq \yy ) | \mathcal{P} \}$ w.r.t. to the projections $\Pp, \Qq$ and then fixing $\Pp, \Qq$ and minimizing $L(\aaa,\bbb)$ w.r.t. the thresholds $\aaa, \bbb$. 
%
Since the problem is separable (as we have seen, projection in each dimension corresponds to a singular vector of the positives covariance matrix), we consider for simplicity one-dimensional projections. % and the related loss, $\EE \{ (\ppp^\Tr \xx ) (\qqq^\Tr \yy ) | \mathcal{P} \}$.

The whole method is summarized in Algorithm~\ref{alg:2}. Since it considers (though implicitly) the intra-modal dissimilarities in addition to the inter-modal dissimilarity, we refer to it as {\em multimodal kernel diff-hash} (MM-kDIF).

\subsection{Projection computation}

Let $k_X : X\times X \rightarrow \RR$ be a positive semi-definite kernel, and let $\phi: \xx \mapsto k_X(\cdot, \xx)$. The map $\phi$ maps the data into some feature space, which we represent here as a Hilbert space $\mathcal{V}$ (possibly of infinite dimension) with an inner  product $\langle \cdot, \cdot \rangle_\mathcal{V}$. It satisfies $k_X(\xx,\xx') = \langle k_X(\cdot, \xx), k_X(\cdot, \xx') \rangle_{\mathcal{V}} = \langle \phi(\xx), \phi(\xx') \rangle_{\mathcal{V}}$. 
Same way, we define the kernel $k_Y : Y\times Y \rightarrow \RR$ and the associated map $\psi: \yy \mapsto k_Y(\cdot, \yy)$ to some other Hilbert space $(\mathcal{V}', \langle \cdot, \cdot \rangle_{\mathcal{V}'})$ for the second modality. 

The idea of kernelization is to replace the original data $X,Y$ with the corresponding feature vectors $\phi(X), \psi(Y)$, replacing the linear projections $\ppp^\Tr \xx$ and $\qqq^\Tr \yy$ with 
\begin{eqnarray*}
p(\xx) &=& \sum_{i=1}^l \alpha_i \langle \phi(\xx_i), \phi(\xx) \rangle_{\mathcal{V}} =  \aalpha^\Tr [ k_X(\xx_1, \xx) \hdots k_X(\xx_l, \xx)] \\
q(\yy) &=& \sum_{j=1}^{l'} \beta_j \langle \psi(\yy_j), \psi(\yy) \rangle_{\mathcal{V}'} = \bbeta^\Tr [k_Y(\yy_1, \yy) \hdots k_Y(\yy_{l'}, \yy)],
\end{eqnarray*}
respectively. Here, $\aalpha, \bbeta$ are sunknown linear combination coefficients, and $\xx_1,\hdots, \xx_l$ and $\yy_1,\hdots, \yy_{l'}$ denote some representative points of each modality acting as respective bases of subspaces used for the representation of data in each modality.  

In this formulation, the approximate loss becomes 
\begin{eqnarray*}
\hat{L}(\aalpha, \bbeta) &=& \frac{1}{|\mathcal{N}|} \sum_{(\xx,\yy)\in \mathcal{N}} p(\xx)  q(\yy) -\frac{\gamma}{|\mathcal{P}|} \sum_{(\xx,\yy)\in \mathcal{P}} p(\xx)  q(\yy) \\
&=& \frac{1}{|\mathcal{N}|} \sum_{(\xx,\yy)\in \mathcal{N}}  \sum_{i=1}^l \alpha_i \langle \phi(\xx_i), \phi(\xx) \rangle_{\mathcal{V}} \sum_{j=1}^{l'} \beta_j \langle \psi(\yy_j), \psi(\yy) \rangle_{\mathcal{V}'} \\
&-&\frac{1}{|\mathcal{P}|} \sum_{(\xx,\yy)\in \mathcal{P}}  \sum_{i=1}^l \alpha_i \langle \phi(\xx_i), \phi(\xx) \rangle_{\mathcal{V}} \sum_{j=1}^{l'} \beta_j \langle \psi(\yy_j), \psi(\yy) \rangle_{\mathcal{V}'} \\
&=& \frac{1}{|\mathcal{N}|} \sum_{(\xx,\yy)\in \mathcal{N}}  \sum_{i=1}^l \alpha_i k_X(\xx_i, \xx) \sum_{j=1}^{l'} \beta_j k_Y(\yy_j, \yy) \\
&-&\frac{\gamma}{|\mathcal{P}|} \sum_{(\xx,\yy)\in \mathcal{P}}  \sum_{i=1}^l \alpha_i k_X(\xx_i, \xx) \sum_{j=1}^{l'} \beta_j k_Y(\yy_j, \yy) \\
&=&  \frac{1}{|\mathcal{N}|} \aalpha^\Tr \Kk^\mathcal{N}_X  (\Kk^\mathcal{N}_Y)^\Tr \bbeta - \frac{\gamma}{|\mathcal{P}|} \aalpha^\Tr \Kk^\mathcal{P}_X  (\Kk^\mathcal{P}_Y)^\Tr \bbeta, 
\end{eqnarray*}
where $\Kk^\mathcal{P}_X$ and $\Kk^\mathcal{P}_Y$ denote $l\times |\mathcal{P}|$ and $l'\times |\mathcal{P}|$ matrices, and $\Kk^\mathcal{N}_X$ and $\Kk^\mathcal{N}_Y$ denote $l\times |\mathcal{N}|$ and $l'\times |\mathcal{N}|$ matrices with elements $k_X(\xx_i,\xx)$ and $k_Y(\yy_i,\yy)$, respectively. 
The optimal projection coefficients $\aalpha, \bbeta$ minimizing $L$ are given as the largest left and right singular vectors of the $l\times l'$ matrix $$\Kk = \frac{1}{|\mathcal{N}|} \Kk^\mathcal{N}_X  (\Kk^\mathcal{N}_Y)^\Tr  - \frac{\gamma}{|\mathcal{P}|}  \Kk^\mathcal{P}_X  (\Kk^\mathcal{P}_Y)^\Tr.$$

The kernels $k_X, k_Y$ can be selected in a way to incorporate the intra-modal similarities which are not accounted for in the previously discussed cross-modality hashing problem. For example, a classical choose is the Gaussian kernel, $k_X(\xx,\xx') = e^{-d^2_X(\xx,\xx')}$ and $k_Y(\yy,\yy') = e^{-d^2_Y(\yy,\yy')}$. This way, we account both for the inter-modal similarity $d_{XY}$ (through the definition of the positive set $\mathcal{P}$) and the intra-modal similarities $d_X, d_Y$ (through the definition of the kernels $d_X, d_Y$). 
Furthermore, the dimensionality of the hash is now bounded by the number of the basis vectors,  $m  \leq \min\{l,l' \}$, which can be arbitrary and in practice limited only by the training set size and computational complexity. 
Finally, the use of kernels generalizes the embeddings to be of a more generic rather than affine form.

\subsection{Threshold selection}

As previously, the threshold should be selected to minimize the false positive and negative rates for each dimension of the projection,   
\begin{eqnarray}
\label{eq:fn2}
\mathrm{FN}(a,b) &=& 
\mathrm{Pr}(p(\xx)  < -a | \mathcal{P}) \cdot (1 - \mathrm{Pr}( q(\yy) < -b | \mathcal{P})) \nonumber \\
&+& \mathrm{Pr}(q(\yy)  < -b | \mathcal{P}) \cdot (1 - \mathrm{Pr}( p(\xx) < -a | \mathcal{P})); \\
\label{eq:fp2}
\mathrm{FP}(a,b) &=& 
\mathrm{Pr}(p(\xx)  < -a | \mathcal{N}) \cdot \mathrm{Pr}( q(\yy) < -b | \mathcal{N}) \nonumber \\
&+& \mathrm{Pr}(q(\yy)  < -b | \mathcal{N}) \cdot \mathrm{Pr}( p(\xx) < -a | \mathcal{N}).
\end{eqnarray}
The optimal thresholds are obtained as 
\begin{eqnarray}
\label{eq:opt_thr2}
(a^*,b^*) &=& \mathop{\mathrm{argmin}}_{a,b} \gamma\mathrm{FN}(a,b) + \mathrm{FP}(a,b).
\end{eqnarray}

\begin{algorithm}[t!]
%\linesnumbered
\KwIn{Positive and negative sets $\mathcal{P}, \mathcal{N} \subset X\times Y$ of multimodal data of dimensionality $n$ and $n'$; 
Dimensionality of the hash $m$; Kernels $k_X, k_Y$; Bases $\xx_1,\hdots,\xx_l$ and $\yy_1,\hdots,\yy_{l'}$.}
\KwOut{Optimal combination coefficient matrices $\Aa, \Bb$ of size $m\times l, m\times l'$; optimal offset vectors $\aaa, \bbb$ of size $m\times 1$.}

Compute the kernel matrices $\Kk^\mathrm{P}_{X}, \Kk^\mathrm{P}_{Y}, \Kk^\mathrm{N}_{X}, \Kk^\mathrm{N}_{Y}$.

Perform singular value decomposition  $$\frac{1}{|\mathcal{N}|} \Kk^\mathcal{N}_X  (\Kk^\mathcal{N}_Y)^\Tr  - \frac{\gamma}{|\mathcal{P}|}  \Kk^\mathcal{P}_X  (\Kk^\mathcal{P}_Y)^\Tr  = \Uu \Ss \Vv^\Tr.$$ 

\For{$i=1,\dots,m$}{

Set the $i$th row of the coefficient matrices to be the $i$th largest left and right singular vectors, $\aalpha^\Tr_i = \uu^\Tr_{n-i+1}$ and $\bbeta^\Tr_i = \vv_{n'-i+1}^\Tr$.

Compute the projections $p_i(\xx) = \aalpha_i^\Tr (k_X(\xx_1,\xx),\hdots,k_X(\xx_l,\xx))$, $q_i(\yy) = \bbeta_i^\Tr (k_Y(\yy_1,\yy),\hdots,k_Y(\yy_{l'},\yy))$.

Compute the probabilities $\mathrm{Pr}(p_i(\xx) < \cdot | \mathcal{P})$, $\mathrm{Pr}(p_i(\yy) < \cdot  | \mathcal{P})$, 
 $\mathrm{Pr}(p_i(\xx) < \cdot | \mathcal{N})$, $\mathrm{Pr}(p_i(\yy)< \cdot  | \mathcal{N})$. 

Compute the rates $\mathrm{FP}_i(a_i,b_i), \mathrm{FN}_i(a_i,b_i)$, as a function of the thresholds $a_i, b_i$ according to (\ref{eq:fn2})--(\ref{eq:fp2}).

Compute the optimal thresholds 
\begin{eqnarray*}
(a_i^*,b_i^*) &=& \mathop{\mathrm{argmin}}_{a,b} \gamma \mathrm{FN}_i(a,b) + \mathrm{FP}_i(a,b).
\end{eqnarray*}

}

\caption{Multimodal kernel diff-hash algorithm (MM-kDIF). \label{alg:2}}
\end{algorithm}

\subsection{Hash function application}

Once the linear combination coefficients $\Aa, \Bb$ and thresholds $\aaa,\bbb$ are computed, given new data points $\xx, \yy$, we construct the corresponding $m$-dimensional binary hash vectors as $\xi(\xx) = \sign(\Aa (k_X(\xx_1,\xx),\hdots,k_X(\xx_l,\xx))^\Tr + \aaa)$ and $\eta(\yy) = \sign(\Bb (k_Y(\yy_1,\yy),\hdots,k_Y(\yy_{l'},\yy))^\Tr + \bbb)$.

%by first computing $\phi(\xx)[\xx_1,\hdots,\xx_l] = (k_X(\xx_1,\xx),\hdots,k_X(\xx_l,\xx))$ and $\psi(\yy)[\yy_1,\hdots,\yy_{l'}] = (k_Y(\yy_1,\yy),\hdots,k_Y(\yy_{l'},\yy))$, and then computing $\xi(\xx) = \sign(\Aa \phi(\xx)[\xx_1,\hdots,\xx_l] + \aaa)$ and $\eta(\yy) = \sign(\Bb \psi(\yy)[\yy_1,\hdots,\yy_{l'}] + \bbb)$.

\section{Results}

%\subsection{Synthetic data}

%In the first experiment, 
To test the performance of the algorithms, 
we created simulated multimodal data of dimensionality $n=128$  and $n'=64$. 
In each modality, the data was created as follows: first, $K = 25, 50$, and $100$ random vectors were generated as ``centers''. To each ``center'' ($128$- or $64$-dimensional, respectively), i.i.d. Gaussian noise with different standard deviation in each dimension (varying between $3-6$) was added. 
Binary inter-modal similarity partitioned the dataset into $K$ classes. 
As the intra-modal dissimilarity in each modality, we used the Mahalanobis metric with respective diagonal covariance matrix.

We compared boosting-based CM-SSH \cite{cmssh} and our CM-DIF and MM-kDIF methods. 
Hash of different dimension $m$ was used for CM-SSH and MM-kDIF; for CM-DIF was used. 
We used tradeoff parameter $\gamma = 10$. For MM-kDIFF, we used bases of size $l=l'=10^3$ and Gaussian kernels of the form 
\begin{eqnarray*}
k_X(\xx,\xx') &=& e^{ -d^2_X(\xx,\xx') } =  e^{ -(\xx - \xx')^\Tr \SSigma_X^{-1/2}(\xx - \xx') }; \\
k_Y(\yy,\yy') &=& e^{ -d^2_Y(\yy,\yy') } =  e^{ -(\yy - \yy')^\Tr \SSigma_Y^{-1/2}(\yy - \yy') }. 
\end{eqnarray*}
For CM-SSH, the settings were according to \cite{cmssh}.

The training set consisted of $10^4$ positive and $10^5$ negative pairs. 
The training time for $m=50$ was approximately $162$, $0.62$, and $28$ seconds for CM-SSH, CM-DIF, and MM-kDIF, respectively. 
Testing was performed on a set of $5\times 10^3$ pairs, using data from one modality as a query and data from another modality as the database. Performance was measured as mean average precision (mAP) and equal error rate (EER). Ideal performance is $\mathrm{mAP}=1$ and $\mathrm{EER}=0$.   
%
%
%Table~\ref{tab1} compares the training time of different algorithms. Our method have a clear advantage. 

\begin{figure}[tpb]
    \begin{center}
\includegraphics[width=0.49\linewidth,bb=25 5 415 337,clip]{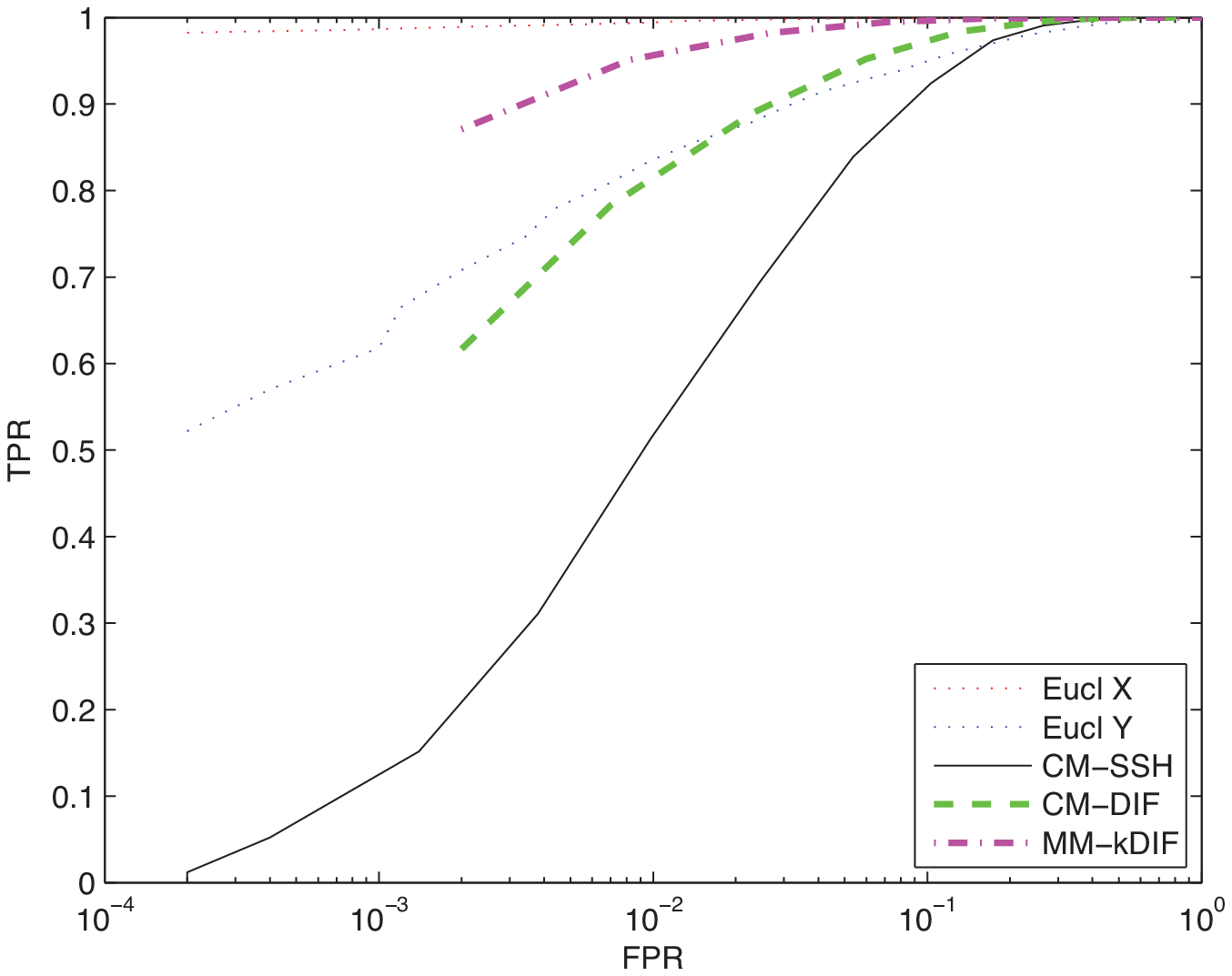} \vspace{2mm}
\includegraphics[width=0.49\linewidth,bb=25 5 415 337,clip]{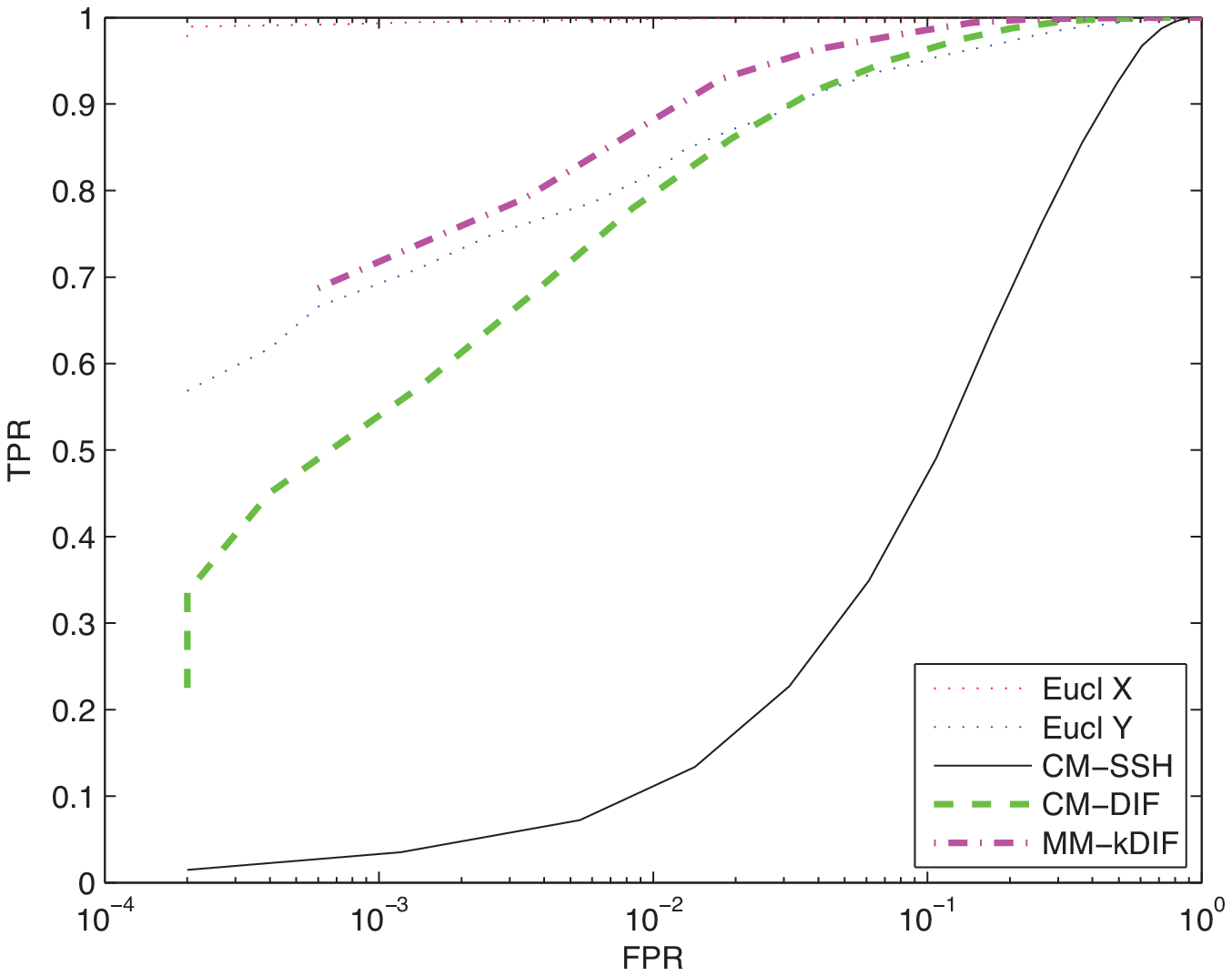}\\
\includegraphics[width=0.49\linewidth,bb=25 5 415 337,clip]{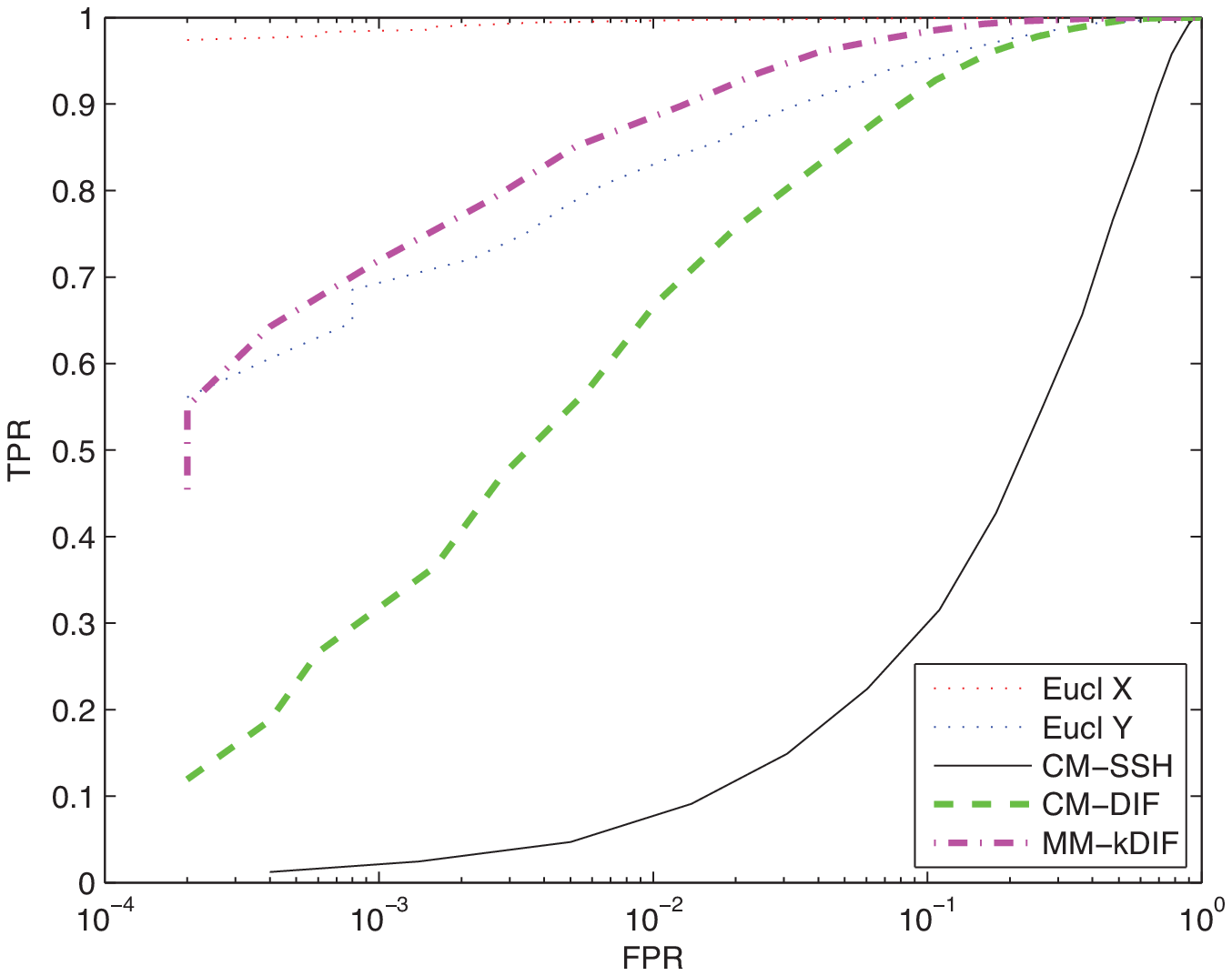}\
   \caption{\label{fig:roc} \small ROC curves showing the performance of different multimodal hashing algorithms on synthetic data retrieval experiment with $K=25, 50, 100$ (ordered left-to-right, top-to-bottom) classes. Hash length used is $m=25, 50$, and $100$, respectively (in the last case, $m=64$ is used for CM-DIF). For comparison, unimodal retrieval in each modality using Euclidean distance is shown (dotted). }
    \end{center}
\end{figure}

\begin{figure}[tpb]
    \begin{center}
\includegraphics[width=0.49\linewidth,bb=20 10 415 337,clip]{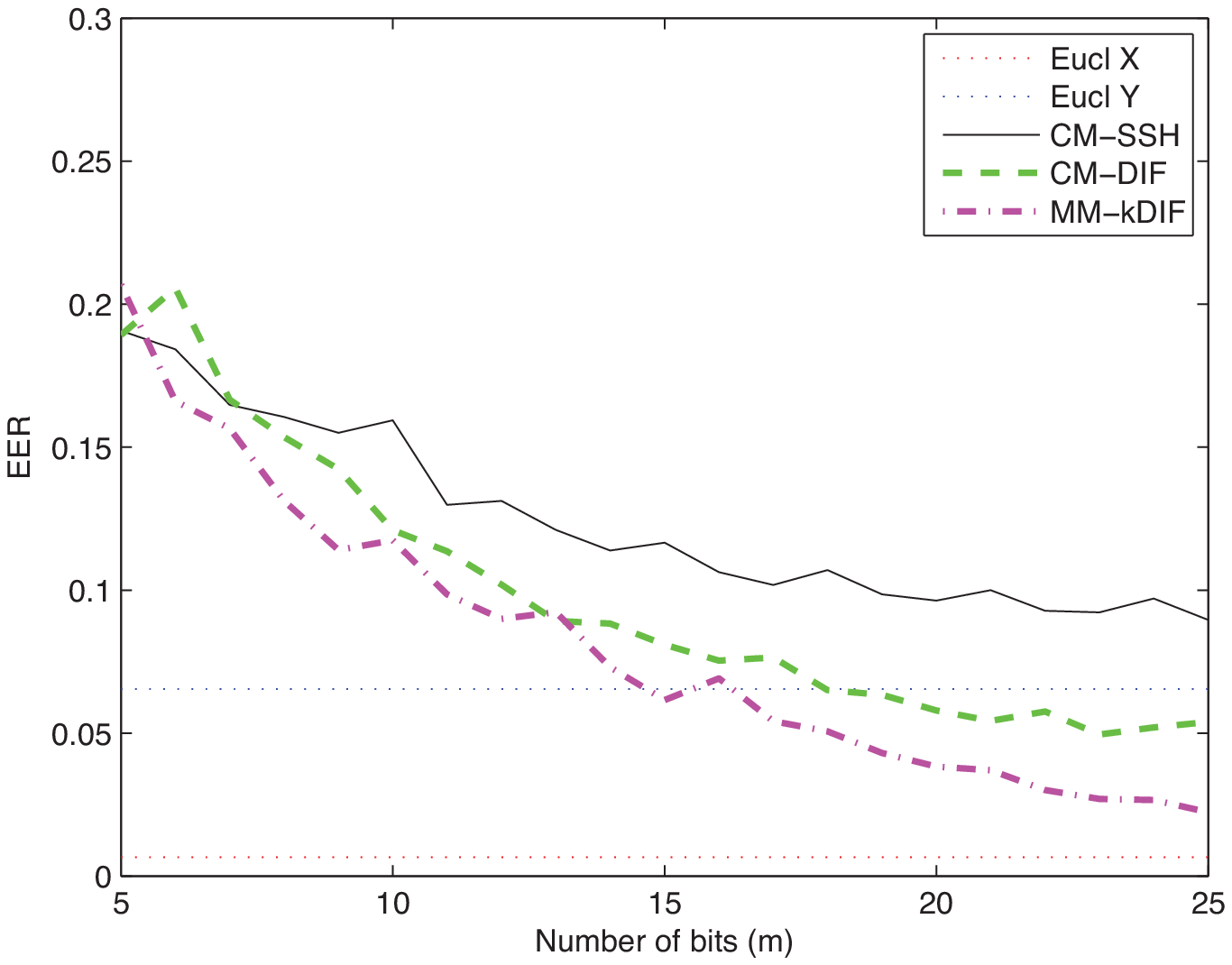} \vspace{2mm}
\includegraphics[width=0.49\linewidth,bb=20 10 415 337,clip]{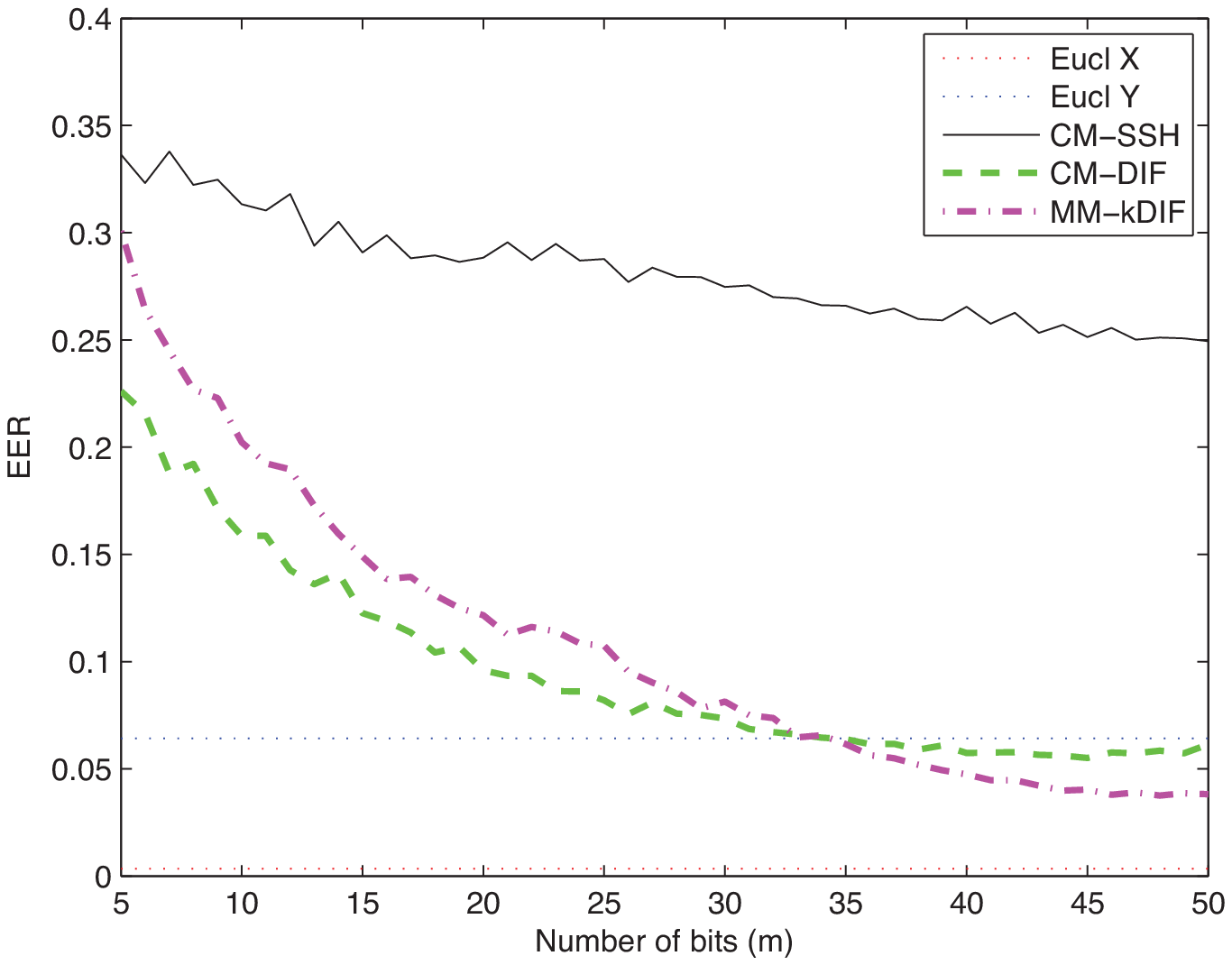}\\
\includegraphics[width=0.49\linewidth,bb=20 10 415 337,clip]{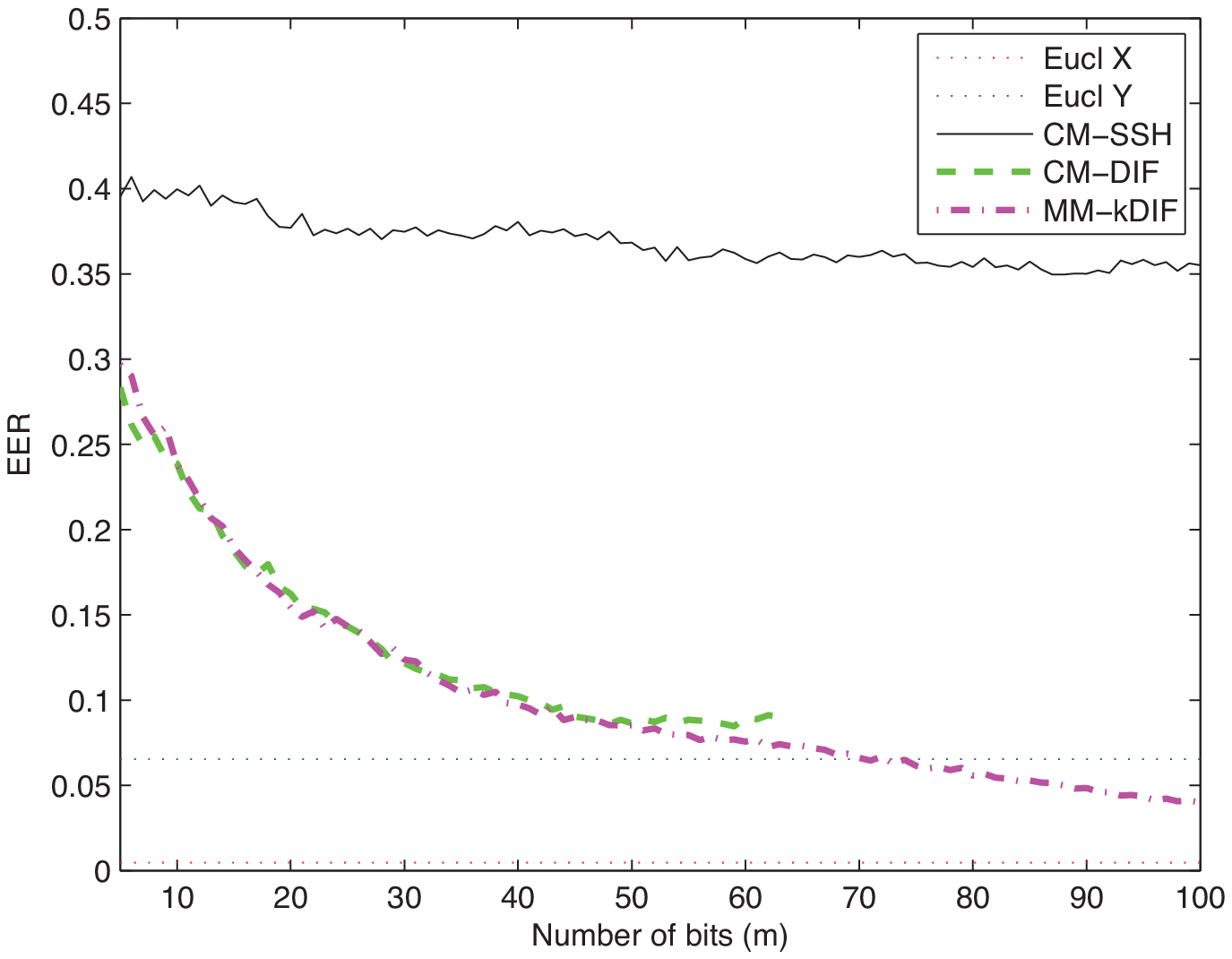}\
   \caption{\label{fig:eer} \small Performance (EER) of different multimodal hashing algorithms on synthetic data retrieval experiment with $K=25, 50, 100$ (ordered left-to-right, top-to-bottom) classes as a function of the hash length $m$. For comparison, unimodal retrieval in each modality using Euclidean distance is shown (dotted). In the last case, the length of CM-DIF hash is limited by the data dimensionality.}
    \end{center}
\end{figure}

Figures~\ref{fig:roc}--\ref{fig:eer} show the performance of different multimodal hashing algorithms as a function of $m$ for datasets with a different number of classes. For comparison, we show the performance of unimodal retrieval (Euclidean distance). Our methods clearly outperform CM-SSH both in accuracy and training time. Moreover, the performance of CM-SSH seems to fall dramatically with increasing complexity of the dataset (more classes), while our methods continue producing good performance.
%
%The performance is summarized in Table~\ref{tab1}. 

%
%\begin{table}[tp]
%   \caption{ \small Training time (in sec) of different multimodal hashing algorithms using training sets of equal size and hash of equal length.  }
%   \center
%\begin{tabular}{lcccccc}
%%\hline\hline
%\bf Dataset  & $|\mathcal{P}|$ & $|\mathcal{N}|$ & \hspace{2mm} $m$ \hspace{2mm}  & \bf CM-SSH & \bf CM-DIF  & \bf MM-kDIF    \\
%\hline
%Random-25		& $10^4$ & $10^5$ & $60$ &		 $1311$ & $162$   &  $20.6$ \\
%Random-50		& $10^4$ & $10^5$ & $60$ &		 $838$   & $142$   &  $20.5$ \\
% \hline
%\end{tabular}
%\label{tab1}
%\end{table}

\bibliographystyle{plain}
\bibliography{ldahash}

\end{document}